\pdfoutput=1

\documentclass[11pt]{article}

\usepackage[]{ACL2023}
\usepackage{times}
\usepackage{latexsym}

\usepackage[T1]{fontenc}

\usepackage[utf8]{inputenc}

\usepackage{microtype}

\usepackage{inconsolata}

\usepackage{multirow}  
\usepackage{amssymb}   
\usepackage{graphicx}  
\usepackage{amsmath}
\usepackage{subfigure} 
\usepackage{makecell} 
\usepackage{booktabs} 
\usepackage{enumitem} 
\usepackage{multirow} 
\usepackage{booktabs} 
\usepackage[section]{placeins} 
\usepackage{amsmath}
\usepackage{wrapfig}
\usepackage{graphicx}

\newcommand{\exper}[1]{\textsc{#1}}
\newcommand{\expmb}[1]{\mathbf{#1}}
\newcommand{\expobj}[1]{\mathcal{L}_\text{#1}}

\newcommand{\ww}{\expmb{w}}

\newcommand{\cx}[1]{\mathbf{x}_{#1}}
\newcommand{\ptheta}{p_\theta}

\newcommand{\qphi}{q_\phi}
\newcommand{\Lsimple}{\expobj{simple}}
\newcommand{\Lvlb}{\expobj{vlb}}

\newcommand{\Emb}{\exper{Emb}}

\newcommand{\model}[0]{DiffCap~}

\title{DiffCap: Exploring Continuous Diffusion on Image Captioning}

\author{
  Yufeng He\footnotemark[1] \\
  Peking University \\
  yufeng.he@stu.pku.edu.cn \\ \And
  Zefan Cai\footnotemark[1] \\
  Peking University \\
  zefncai@gmail.com\\ \AND
  Xu Gan \\
  Fudan University \\
  ganx21@m.fudan.edu.cn \\ \And
  Baobao Chang\footnotemark[2] \\
  Peking University \\
  chbb@pku.edu.cn \\
  }

\begin{document}

\maketitle
\begin{abstract}

Current image captioning works usually focus on generating descriptions in
an autoregressive manner. However, there are limited works that
focus on generating descriptions non-autoregressively, which brings more decoding diversity.
Inspired by the success of diffusion models on generating natural-looking images, 
we propose a novel method DiffCap to apply continuous diffusions on image captioning. 
Unlike image generation where the output is fixed-size and continuous,
image description length varies with discrete tokens. Our method transforms
discrete tokens in a natural way and applies continuous diffusion on them to 
successfully fuse extracted image features for diffusion caption generation. 
Our experiments on COCO dataset demonstrate that our method uses a much simpler structure
to achieve comparable results to the previous non-autoregressive works.
Apart from quality, an intriguing property of DiffCap is its high diversity during generation, 
which is missing from many autoregressive models. We believe our
method on fusing multimodal features in diffusion language generation will
inspire more researches on multimodal language generation tasks for its simplicity and decoding flexibility.\footnote{https://github.com/arealgoodname/diffcap}

\end{abstract}

\section{Introduction}

\def\thefootnote{*}\footnotetext{equal contribution}\def\thefootnote{\arabic{footnote}}
\def\thefootnote{\dag}\footnotetext{corresponding author}\def\thefootnote{\arabic{footnote}}

Image captioning has been studied for years, it takes images input as condition
and expects outputs as description texts of corresponding images.
Through years of research, the text generation decoder can be classified 
into two general classes, i.e., autoregressive and non-autoregressive. 
Most of the previous works are autoregressive models based on the encoder-decoder 
architecture\cite{2014Deep}\cite{2015Show}\cite{2020Stack}
, which suffers from causing sequential error accumulation\cite{gao2019masked} and lacking generation diversity. 
In contrast, non-autoregressive models predict tokens in parallel,
which alleviate such problems.

As non-autoregressive generation models, diffusion models\cite{2020Denoising} has impressed the 
community with its generation quality and diversity on many generation
tasks including imagegeneration\cite{2021Cascaded}\cite{2021Diffusion}
image super-resolution\cite{2021Image} , and audio generation \cite{2020DiffWave}. 
Since most NLP tasks are generative or can be transferred into generation problems,
a natural idea is to apply diffusion model on these tasks. However, 
due to the discrete nature of languages, simply applying diffusion models to language generation is quite challenging. There are series of methods 
proposed to model discrete data, where \cite{Chen2022AnalogBG} represent the discrete data as binary bits,  Diffusion-LM\cite{Li2022DiffusionLMIC} 
trained an embedding function to transfer discrete token to continuous latent representations, but when generating it tends to generate many [UNK] tokens which degrades the generation quality. 

In this paper, we propose a novel diffusion image captioning method named DiffCap, which is non-autoregressive based on the continuous diffusion. We aim to improve the generation diversity with diffusion models, and to solve the gap between discrete tokens and continuous diffusion process, we trained an embedding function to project tokens to continuous representations, used KNN for reverting back to discrete tokens.\\
 Our main contributions are as follows:
 
    1) We propose a non-autoregressive image captioning method based on continuous diffusion, proved that it generates much more diverse captions than autoregressive models, and comparable to the previous non-autoregressive models with our model structure being much simpler.
    
    2) Our method successfully fuses visual features with diffusion language generation, it is simple and can be easily integrated into other diffusion
    based multi-modal language generation tasks, this will benefit the future research on multi-modal language generation.
    
To the best of our knowledge,
this is the first work to apply continuous diffusion to image captioning task.

\section{Related work}
\subsection{Image captioning}
Image captioning aims to generate syntactically and semantically correct sentences to describe images. A large number of deep learning-based techniques have been proposed for this task, which usually use an encoder-decoder architecture\cite{2014Deep}\cite{2015Deep}\cite{2015Show}. To solve the problem of information loss, \cite{2016Self} improves the image features' quality by employing more fine-grained features;\cite{2017Cascade} adopts a cascade network, which can exploit the deep semantic contexts contained in the image. Later, attention mechanisms are used for better caption generation: \cite{2017Knowing} used attention mechanisms for fusion between image feature and text feature;\cite{Le2021GloballocalAF} combined local and global features from images;\cite{2020Stack} propose an innovative multi-stage architecture to handle both visual-level and semantic-level information of an input image. In addition, some deep generative model frameworks have been applied to image captioning, such as generative adversarial network\cite{Dai2017TowardsDA}\cite{Feng2018UnsupervisedIC}\cite{2020MSCap}\cite{Chen2018GroupCapGI}and variational auto-encoder\cite{Kim2019VariationalAM}. Besides, a lot of methods tried to improve the generation quality: \cite{Liu2016ImprovedIC}improved metrics using reinforcement learning;\cite{He2017ImageCG} used POS tagging to help the generation. Owing to the success of large-scale pretrained model, \cite{Mokady2021ClipCapCP}make use of CLIP feature;\cite{Li2022mPLUGEA} introduced a new visual language architecture with skipping connections between transformer layers to address the information asymmetry between vision and language modality.\cite{Nguyen2022GRITFA} proposes a Transformer-only neural architecture to fuse grid-based features and region-based features.

However, those autoregressive models suffer from issues such as 
sequential error accumulation and a lack of decoding diversity. 
Multiple non-autoregressive models\cite{Gao2019MaskedNI}\cite{Fei2021PartiallyNI} and 
Semi-autoregressive\cite{Yan2021SemiAutoregressiveIC} models are proposed to 
address these issues. 

\subsection{Diffusion Model}
Existing generative models like GAN\cite{2016NIPS} and VAE 
\cite{Kingma2013AutoEncodingVB}have problems such as training instability, 
mode collapsing. While solving these problems, diffusion models have 
state-of-the-art sample quality in many tasks\cite{2021Cascaded}
\cite{2021Diffusion}\cite{Nichol2021GLIDETP}.\cite{2020Denoising} 
proposed a parameterized Markov chain trained by variational inference,
and \cite{Nichol2021ImprovedDD} improved the log-likelihood. 
In image generation and audio generation where data is naturally 
in continuous form, diffusion models achieved outstanding performance.
But it is not diffusion model's nature to deal with discrete data ,
some works have been proposed to tackle this problem,
\cite{Hoogeboom2021AutoregressiveDM}
introduced a new model at the intersection of autoregressive models and 
discrete diffusion models.\cite{Chen2022AnalogBG} represented the discrete 
data as binary bits and used a continuous diffusion model to model these bits.
Furthermore, \cite{Li2022DiffusionLMIC}transferred tokens to continuous embedding
representations and modeled them with continuous diffusion models which our work is similar to.

\subsection{CLIP}
CLIP \cite{Radford2021LearningTV}is a vision-language pretrained model 
based on contrastive learning that consists of a text encoder and an image encoder. 
When training, paired image-text embeddings are pushed together in a same semantic space,
while un-paired image-text embeddings are pushed away,
this mechanism with large scale pretraining gives CLIP strong performance on encoding image and text features. 
Clip used ViT\cite{Dosovitskiy2020AnII} as visual encoder, which is proved to be an excellent image feature extractor.

\section{Background}
\label{sec:background}


We aim to apply continuous diffusion models to image captioning problems. To setup the context, we will review some of the basics of continuous diffusion models.

\subsection{Diffusion Models}
\label{ssec:diffution_models}

Given data distribution $R^d$, a diffusion model is a latent variable model that models the data $\cx{0} \in R^d$ as a Markov chain $\cx{T}, \cdots, \cx{0}$ where $\cx{T}$ is a pure Gaussian noise and $\cx{0}$ is a variable from $R^d$. $\cx{t}$ represents the intermediate representation in the diffusion generation stages. 

The training of the continuous diffusion model includes construction of noised samples as forward process, and denoising Gaussian noise back to the original distribution as backward process.
In forward process, the sequence of continuous latent variables $\cx{1:T}$ is constructed by incrementally adding Gaussian noise to data $\cx{0}$ to generate noised samples $[\cx{1}, \cdots, \cx{T}]$. At final diffusion step $T$, samples $\cx{T}$ are approximately pure Gaussian noise. 
Each transition $\cx{t-1} \rightarrow \cx{t}$ is parametrized by $q(\cx{t} \mid \cx{t-1}) = \mathcal{N} (\cx{t} ; \sqrt{1-\beta_t} \cx{t-1}, \beta_t \mathbf{I})$, where the hyperparameter $\beta_t$ is the amount of noise added at diffusion step $t$.  
In backward process, the diffusion model is trained to iteratively denoise the sequence of latent variables $\cx{T:1}$ to approximate samples of the original distribution. 
Each denoising transition $\cx{t} \rightarrow \cx{t-1}$ is parametrized by the model $\ptheta(\cx{t-1}\mid \cx{t}) = \mathcal{N}(\cx{t-1}; \mu_\theta(\cx{t}, t), \Sigma_\theta(\cx{t}, t))$.  

The diffusion model is trained to maximize the marginal likelihood of the data $\mathbb{E}_{\cx{0} \sim p_\text{data} }\log p_\theta(\cx{0})$, and the canonical objective is the variational lower bound of $\log p_\theta(\cx{0})$ \cite{sohl2015deep}:

\begin{equation}
    L_T = \log \frac{q(\cx{T} | \cx{0})}{\ptheta(\cx{T})}
\end{equation}

\begin{equation}
    L_t = \sum_{t=2}^T \log \frac{q(\cx{t-1} | \cx{0},\cx{t})} {\ptheta(\cx{t-1} | \cx{t}) }
\end{equation}

\begin{equation}
    L_0 = \log \ptheta( \cx{0} | \cx{1})
\end{equation}

\begin{equation}
\Lvlb (\cx{0}) = \mathop{\mathbb{E}}_{q(\cx{1:T} | \cx{0})} \left[L_T + L_t + L_0)\right]
\label{equation:lvlb}
\end{equation}


However, this objective can be unstable and requires many optimization tricks to stabilize \cite{nichol2021improved}. To circumvent this issue, \citep{ho2020denoising} devised a simple surrogate objective that expands and reweights each KL-divergence term in $\Lvlb$ to obtain a mean-squared error loss which we will refer to as:

\begin{equation}
\label{equation:lsimple}
\Lsimple (\cx{0}) = \sum_{t=1}^T \mathop{\mathbb{E}}_{q(\cx{t} \mid \cx{0})} || \mu_\theta(\cx{t}, t) - \hat{\mu}(\cx{t}, \cx{0}) ||^2
\end{equation}

where $\hat{\mu}(\cx{t}, \cx{0})$ is the mean of the posterior $q(\cx{t-1} | \cx{0},\cx{t})$ which is a closed from Gaussian, and $\mu_\theta(\cx{t}, t)$ is the predicted mean of $\ptheta(\cx{t-1} \mid \cx{t})$ computed by a neural network.

While $\Lsimple$ is no longer a valid lower bound, prior work has found that it empirically made training more stable and improved sample quality.
We will apply similar simplifications for Equation~\eqref{equation:lsimple} in Diffusion-LM~\citep{li2022diffusion} as described in Section~\ref{ssec:diffusion_language} to stabilize training and improve sample quality for caption generation.

\subsection{Diffusion Models for Language Generation}
\label{ssec:diffusion_language}

\citep{li2022diffusion} proposes Diffusion-LM, a new non-autoregressive language model based on continuous diffusion model to iteratively denoise a sequence of Gaussian vectors into word vectors. To apply diffusion models to text generation, they introduced an embedding step and a rounding step to the standard diffusion process, and designed a training objective to jointly learn
the diffusion model’s parameters and word embeddings as an extension of Equation \eqref{equation:lsimple}:

\begin{equation}
 L_{mse} = || \Emb(\ww) - \mu_\theta(\cx{1}, 1) ||^2
\label{equation:loss_x0}
\end{equation}

\begin{equation}
 L_{w} = - \log \ptheta(\ww | \cx{0})
\label{equation:loss_token}
\end{equation}

\begin{equation}
 \mathcal{L}^\text{e2e}_\text{simple} (\ww) = \mathop{\mathbb{E}}_{\qphi(\cx{0:T} | \ww)} \left[\Lsimple(\cx{0}) + L_{x_0} + L_{w} \right]
\label{equation:lsimple_diffusion_lm}
\end{equation}

where $\ptheta(\ww \mid \cx{0}) = \prod_{i=1}^n \ptheta(w_i \mid x_i)$ is a trainable rounding step and  $\ptheta(w_i \mid x_i)$ is a softmax distribution. And $L_{mse}$ and $\Lsimple(\cx{0})$ are both mse loss.

\section{Method}
\label{sec:method}

\begin{figure*}
\centering
\includegraphics[width=\linewidth]{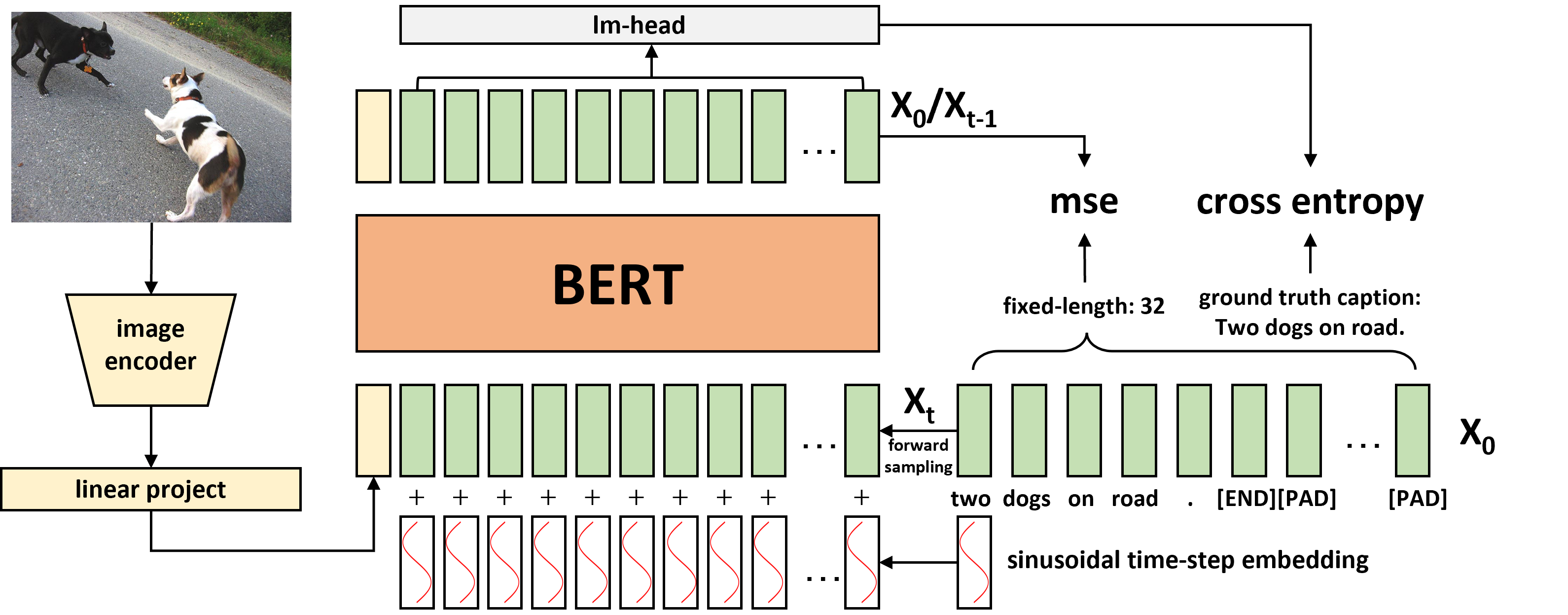}
\caption{
An overview of the architecture of our proposed \model.
}
\label{fig:model-architecture}
\vspace{-10pt}
\end{figure*} 

Our DiffCap model (Fig \ref{fig:model-architecture}) consists of 2 main components: a visual encoder component(left side of figure \ref{fig:model-architecture}) and a diffusion language generation component. The visual encoder component is a pretrained ViT-Base/32 model \citep{Dosovitskiy2020AnII} or a pre-trained CLIP(ViT-B/32) model\citep{Radford2021LearningTV}, which is used to encode the visual input into a fixed-length vector. The diffusion language generation component is a Bert model that fuses representation of the image and diffused language embeddings together to generate the caption. During training, the visual encoder is frozen and the diffusion language generation component is trained end-to-end to perform denoising diffusion. The Bert model was reinitialized since we found Bert pre-trained weights initializing always lead to a lower performance. 

\subsection{Structure}
\label{ssec:}

In visual encoder part, we use output feature at [CLS] token position from ViT-base/CLIP-base to get a 768/512-dimensional image global feature representation. A linear layer is used to project the image global feature to the Bert's hidden dimension. \\
The diffusion language generation component has a similar structure as the Diffusion LM, which is recently proposed as a non-autoregressive language model based on continuous diffusions. It takes a fixed-length sequence of embeddings as input and predicts the denoised embeddings corresponding to the diffusion time-step. The fixed-length sequence comes from padding or truncating ground truth caption. Ground truth caption will be tokenized and padded with [PAD] token, [END] token will be appended to tell the language model a sentence is finished, longer sentence will be truncated without [END] token at the end. We use BPE tokenizer to tokenize the caption, low frequency tokens will be replaced with [UNK] token. These tokens will pass through an embedding layer and form a fixed-size [Batch-size, Fixed-seq-len, Embedding-dim] tensor, which continuous diffusions can be applied to.

\subsection{End-to-end training}
\label{ssec:}

In the forward process, visual encoder first extracts the image feature, these features serve as condition for the diffusion language generation, and will stay the same during the whole forward process, only text embeddings will be diffused. The bert model will take the image feature concatenated with the diffused text embeddings as input, and predict the denoised text embeddings corresponding to the diffusion time-step. When generating the caption, similar rounding function\citep{li2022diffusion} is used to push the denoised text embeddings to the nearest discrete token. The denoise text embeddings will pass through a LM head(linear layer) to get the logits, and then a softmax function is used to get the probability of each token. The token with the highest probability will be selected as the predicted token in the caption. This is similar to the MLM token prediction process in BERT.

We decided to use the same loss function Lsimple as the Diffusion LM to train our model. Which includes a mse loss between the denoised text embeddings and the ground truth text embeddings, and a cross entropy loss between the predicted token and the ground truth token. The mse loss is used to encourage the denoised text embeddings to be close to the ground truth text embeddings X$_{0}$, and the cross entropy loss is used between predicted token from lmhead and ground truth token. The loss function is defined as:

\begin{equation}
    L_\text{final} = \mathcal{L}^\text{e2e}_\text{simple} (\ww) * (\ww != \text{[UNK]})
\end{equation}

Where (w != \text{[UNK]}) as loss mask is for solving the problem of generating [UNK] tokens in Diffusion LM, since when training more than 40\% of tokenized captions will include [UNK] token, as ground truth it encourages the model to predict [UNK] under the guidance of cross entropy loss. Loss mask removes the [UNK] token from the loss calculation, which improved generation performance, the model no longer generates [UNK].

\section{Experiment}
\label{sec:experiment}

\subsection{Dataset}
\label{ssec:dataset}
We choose to test DiffCap on two commonly used image captioning dataset: COCO \citep{coco} and Flickr30k\citep{flickr}, but Flickr30k will be used for ablation study due to our limited computation resources. Flickr30k has about 30k images and 5 captions for every image
We followed Flickr30k's official split to train validate and test our models.
COCO dataset has 123,287 images labeled with 5 captions for each, with 82,783 images in the training set, 40,504 images in the validation set, and 40,775 images in the test set.
We followed the widely used "Karpathy" split\citep{KarpathyF14} in the literature, which splits the COCO dataset into 113,287 images for training, 5,000 images for validation, and 5,000 images for testing.

\subsection{Experiment Setups}
\label{ssec:training_details}

\paragraph{Computational resources} All of our models were trained on a single NVIDIA 3090 GPU with 24GB memory. Every round took less than 12 hours for both Flickr and COCO training. 

\paragraph{Vocabulary} We use BPETokenizer\citep{bpe} to tokenize the captions. Tokens that appear less than 10 times in the training set will be replaced with [UNK] token. We use different vocabulary and embedding dimension for different datasets. For Flickr30k, we use a vocabulary of 5156 tokens and an embedding dimension of 128. For COCO, vocabulary consists of 8016 tokens and embedding dimension is 256.

\paragraph{Time step embedding} We use a sinusoidal embedding\citep{Dosovitskiy2020AnII} to encode the diffusion time-step. When training, time step embedding can be prepending before input sequence or be elementwise added to input sequence. We found these two methods lead to similar generating performance on flickr30k test set, we choose to use prepending since it's easier to implement.

\paragraph{Training Details}
We use CLIP(ViT-B/32) as our visual encoder. We didn't initialize the Bert model or the embedding layer with pre-trained weights since we found it useless. The language generation model together with vocabulary have around 90 million parameters in total, we set initial learning rate as 1e-4 for 50 epochs with batch size = 64, linear learning rate scheduler was used.

\paragraph{Metrics} Following the standard evaluation setup, we report the performances of our model and compare to other methods over five metrics: BLEU@4~\cite{Papineni2002BleuAM}, METEOR~\cite{Banerjee2005METEORAA}, ROUGE-L~\cite{Jain2018TwoCP}, CIDEr-D~\cite{Vedantam2015CIDErCI}, and SPICE~\cite{Anderson2016SPICESP}.

\subsection{Main Results}
\label{ssec: main_results}

\begin{table}[h]
  \centering
  \setlength\tabcolsep{3pt}
  \resizebox{\linewidth}{!}{
  \begin{tabular}{@{}l c c c c c @{}}
    \toprule
    Method & B@4 & C & M & S & R\\
    \midrule
    \multicolumn{6}{l}{\textbf{Auto-regressive models}}\\
    \midrule
    $\rm{G-MLE}$ ~\citep{Dai2017TowardsDA}  & 29.9 & 102.0 & 24.8 & 19.9 & 52.7 \\
        $\rm{ClipCap}$ ~\citep{Mokady2021ClipCapCP}  & 32.1 & 108.3 & 27.1 & 20.1 & - \\
    
    $\rm{DistillVLM}$~\citep{fang2021compressing}  &  35.6 & 120.8 & 28.7 & 22.1 & - \\
    $\rm{ViTCap}$~\citep{fang2022injecting}  & 36.3 & 125.2 & 29.3 &  22.6 & \textbf{58.1} \\
    $\rm{BLIP_L}$~\citep{li2022blip}  & \textbf{40.4} & \textbf{136.7} & - &  - & - \\
    $\rm{GIT_B}$~\citep{wang2022git}  & \textbf{40.4} & 131.4 & \textbf{30.0} & \textbf{23.0} & - \\
    \midrule
    \multicolumn{6}{l}{\textbf{Non-autoregressive models}}\\
    \midrule
        $\rm{MNIC}$~\cite{gao2019masked}   & 31.5 & 108.5 & 27.5 & 21.1& 55.6\\
    $\rm{NAIC_{B,KD}}$~\cite{guo2020non}   & 28.5 & 98.2 & 23.6 & 18.5& 52.3 \\
        $\rm{FNIC_{NAT}}$~\cite{fei2019fast}   & 30.4 & 107.4 & 25.4 & 19.6 & 55.1\\
            $\rm{\model}$ (Ours)  & 31.6 & 104.3 & 26.5 & 19.6 & 57.0 \\
    \bottomrule
  \end{tabular}
  }
  \caption{Performance comparison on COCO captioning Karpathy~\cite{Karpathy2017DeepVA} split with pretraining, where B@4, M, R, C and S and R denote BLEU@4, METEOR, ROUGE-L,
CIDEr and SPICE scores. CIDEr optimization is not used. Our B@4 can be increased to 35.2 after some post-processing techniques, 
such as filtering out repeated words.}
    \label{table:main}
\end{table}

Table \ref{table:main} presents the results of the baseline models and our models on the COCO dataset.
Our DiffCap model shows comparable performance to the previous non-autoregressive models.
Though not as good as the autoregressive models, our model outperforms them in 
generating more diverse captions(see Section \ref{ssec:diversity}). Closest to our work is  \cite{Gao2019MaskedNI},
it uses a BERT model as generator and perform a 2-step refinement
on the generated captions, and a Masked Language Modeling is used to supervise the generation. And, ClipCap \cite{Mokady2021ClipCapCP} uses CLIP
extracted visual feature as prefix to tune GPT-2\citep{radford2019language} model, this looks very similar to our structure.

\subsection{Diversity of Diffusion Generation}
\label{ssec:diversity}

\begin{table*}[h]
\centering
\resizebox{\linewidth}{!}{
\begin{tabular*}{\linewidth}{|l|ll|}
\hline
image             & \multicolumn{2}{l|}{caption}              \\ \hline
\multirow{6}{*}{${\includegraphics[width=0.24\linewidth]{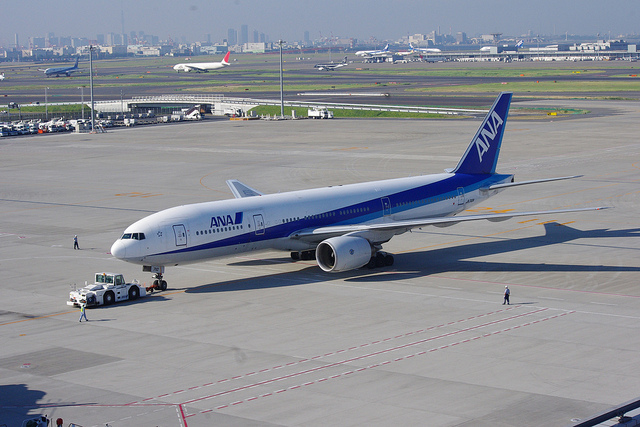}}$} & \multicolumn{1}{l|}{BLIP}                  &  a large jetliner sitting on a tarmac at an airport \\ \cline{2-3} 
                  & \multicolumn{1}{l|}{GIT\_B}                  & a large jetliner sitting on top of an airport tarmac.\\ \cline{2-3} 
                  & \multicolumn{1}{l|}{\multirow{4}{*}{}} &  A large a plane parked on the runway .\\ \cline{3-3} 
                  & \multicolumn{1}{l|}{DiffCap}                  &  A large airplane is waiting down on a tarmac . \\ \cline{3-3} 
                  & \multicolumn{1}{l|}{}                  &  A plane is parked off sitting on a runway .\\ \cline{3-3} 
                  & \multicolumn{1}{l|}{}                  &  A large plane is just at the airport .\\ \hline
\multirow{6}{*}{${\includegraphics[width=0.24\linewidth]{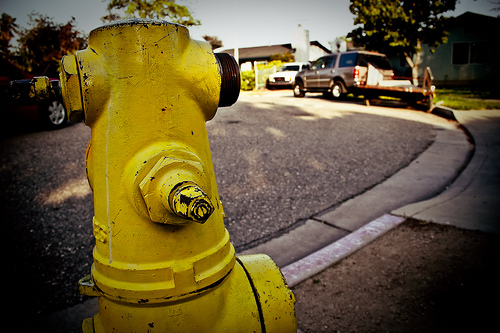}}$} & \multicolumn{1}{l|}{BLIP}                  &   yellow fire hydrant sitting on the side of a road\\ \cline{2-3} 
                  & \multicolumn{1}{l|}{GIT\_B}                  & a fire hydrant on the side of the road.\\ \cline{2-3} 
                  & \multicolumn{1}{l|}{\multirow{4}{*}{}} &  A yellow fire hydrant sits on a  street .\\ \cline{3-3} 
                  & \multicolumn{1}{l|}{DiffCap}                  &  A yellow fire hydrant on the side of the road . \\ \cline{3-3} 
                  & \multicolumn{1}{l|}{}                  &  A yellow fire hydrant parked on the side of the road . \\ \cline{3-3} 
                  & \multicolumn{1}{l|}{}                  &  A yellow fire hydrant in a large city street . \\ \hline
\multirow{6}{*}{${\includegraphics[width=0.24\linewidth]{fables/diversities/baseball.jpg}}$} & \multicolumn{1}{l|}{BLIP}                  &   a man in a baseball uniform is holding a bat and a baseball bat\\ \cline{2-3} 
                  & \multicolumn{1}{l|}{GIT\_B}                  & a baseball player swinging a bat at a ball.\\ \cline{2-3} 
                  & \multicolumn{1}{l|}{\multirow{4}{*}{}} &  A man swinging at a baseball bat at a baseball game .\\ \cline{3-3} 
                  & \multicolumn{1}{l|}{DiffCap}                  &  A baseball player taking a base before a crowd watches. \\ \cline{3-3} 
                  & \multicolumn{1}{l|}{}                  &  A baseball player swinging a bat at home plate . \\ \cline{3-3} 
                  & \multicolumn{1}{l|}{}                  &  A guy who is holding a bat on a baseball field.  \\ \hline
\end{tabular*}}

\caption{captions on COCO Karpathy test set for comparing generation diversity, note that BLIP and GIT-Base always generates the same caption in $8$ times}
\label{table:diversity}
\end{table*}

Table \ref{table:diversity} shows the diversity of different models on COCO test set, we run 4 times for our DiffCap model, 8 times for BLIP and GIT-{Base} model. Result shows that autoregressive models with higher performance always generate the same captions, but our DiffCap has strong performance on generation diversity, captions generated have significant difference between each other.

To better compare the decoding diversity, we provide Inter-Distinct\citep{li-etal-2016-diversity} and Self-Bleu\citep{zhu2018texygen} scores in the Table \ref{table:divmetric}. Inter-Distinct represents the proportion of unique uni-grams and bi-grams in multiple outputs, thus a higher Inter-Distinct score indicates higher decoding diversity. Self-Bleu was calculated by averaging n-gram BLEU score between each pair of generated captions, a lower Self-Bleu score indicates higher diversity. We compare our model with VinVL and BLIP on the test set of MSCOCO, each model generates 5 captions on each image, result shows that our model has made significant improvements in the diversity of the generated results.

\begin{table*}[h]
\centering
\resizebox{\linewidth}{!}{
\begin{tabular}{|l|l|l|l|l|l|l|}
\hline
        & ID uni-gram ↑  & ID bi-gram ↑   & SB uni-gram ↓ & SB bi-gram ↓ & SB tri-gram ↓ & SB 4-gram ↓ \\
\hline
VinVL   & 8.01           & 16.75          & 100           & 100          & 100           & 100         \\
\hline
BLIP    & 6.19           & 14.75          & 100           & 100          & 100           & 99.94       \\
\hline
DiffCap & \textbf{46.61} & \textbf{77.33} & \textbf{71.74}      & \textbf{39.04}        & \textbf{19.78}         & \textbf{10.36}   \\
\hline
\end{tabular}}
\caption{diversity metrics on COCO karpathy test set(5k images)}
\label{table:divmetric}
\end{table*}

\subsection{Visualization and analysis}

\begin{figure}
\centering
\includegraphics[width=\linewidth]{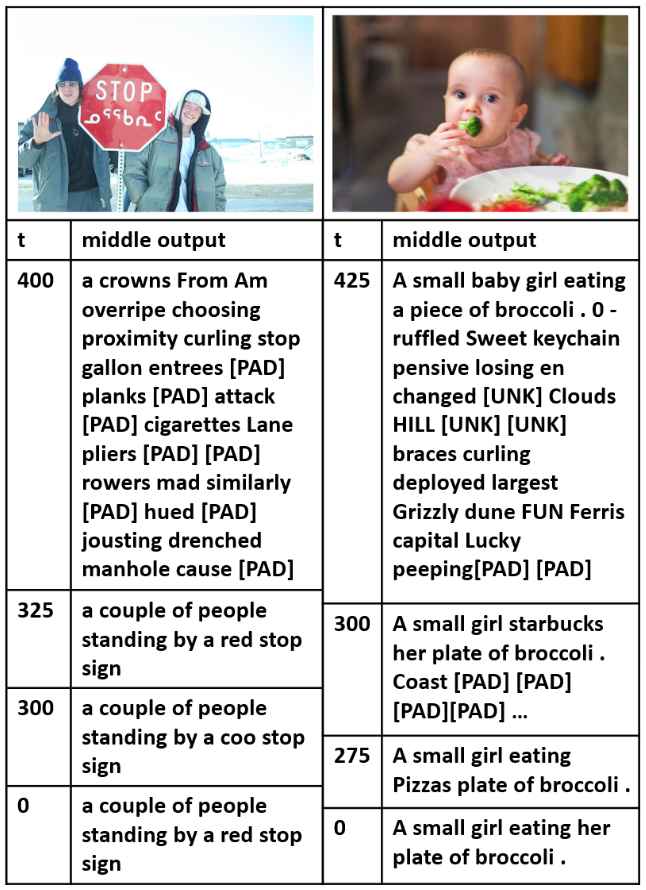}
\caption{middle output visualization}
\label{figure:mid_out}
\end{figure}

Fig \ref{figure:mid_out} shows the denoising middle output of our model on test set,
as expected, the output becomes increasingly clear as the diffusion time-step decreases.
It also shows some interesting phenomenon, in generating caption of the left figure in table 2,
the model first predict 'coo stop' in timestep 300 and then predict 'red stop' in timestep 325,
this shows the gap between applying continuous diffusion to discrete tokens,
where in continuous form like images, images became blurry with diffusing , however,
in language, diffued text embeddings with less noise level does not mean less language distance, e.g., 
'dog' diffused to 'house' then to 'cat' does not mean 'house' has a closer meaning
to 'dog' than 'cat', this hinges the model to understand the diffusion process and perform denoising. Note this gap exists in all other diffusion language generation models, 
we didn't find a way to ease such problem yet, we will leave this to future exploration.

Apart from this, DiffCap shares some weakness with other non-autoregressive models:
generating grammarly incorrect captions and repeating words. These problems can be 
alleviated by using some post-processing techniques, such as filtering out repeated words, but it won't ease the gap between our model and autoregressive models.

\subsection{Ablation Studies}

\paragraph{Fusing Method} We tested 2 commonly used fusing method
prefixing and element-wise adding, tests were done on Flickr30k. Result shows that CLIP image embeddings have better performance than ViT, this may come from the constrastive learning CLIP used. And prefixing embeddings is slightly better than element-wise adding.

\begin{table}[h]
  \centering
  \setlength\tabcolsep{3pt}
  \scalebox{0.86}{
  \begin{tabular}{@{}l c c c @{}}
    \toprule
    Method & B@4 & M & R\\
    \midrule
        CLIP-prefix & \textbf{18.6} & \textbf{15.8} & \textbf{39.3}\\
        CLIP-add & 17.4 & 15.4 & 39\\
        ViT-prefix  & 17.7 & 15.2 & 38.4\\
        ViT-add  & 16.8 & 14.6 & 38.4\\
    \bottomrule
  \end{tabular}
  }
  \caption{ablation results for fusing method}
  \label{tab:fusing}
\end{table}

\paragraph{Beta Scheduler}

Here Table \ref{tab:beta} visualizes noise standard with different beta schedulers, these parameters were used in diffusion process as defined in Sec \ref{ssec:diffution_models}.

Results were listed in Table \ref{tab:beta}, cosine beta scheduler outperforms other schedulers, which is different from Diffusion LM's result, though different beta schedulers have very similar performance.

\begin{table}[h]
\resizebox{\columnwidth}{!}{
\begin{tabular}{lllll}
\hline
\multirow{4}{*}{${\includegraphics[width=0.36\linewidth]{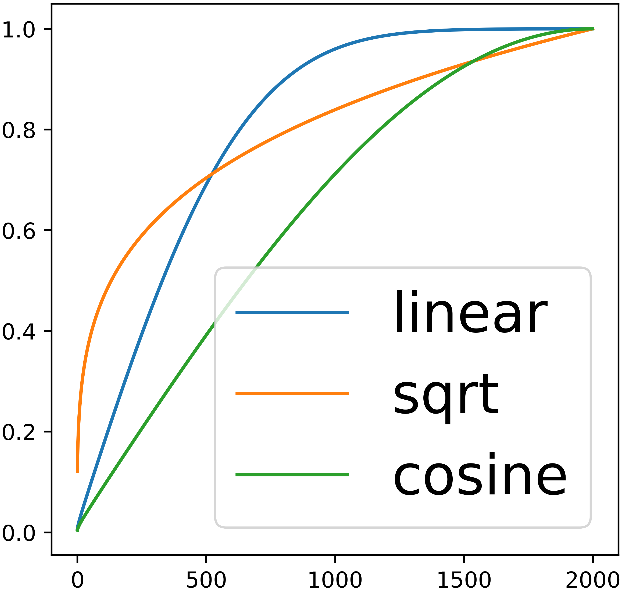}}$} & Method & B@4 & M & R \\ \cline{2-5}
                  & sqrt   & 19.5 & 15.2 & 39.0 \\ [2.8ex]
                  & linear & 18.9 & 16.5 & 39.3 \\  [2.8ex]
                  & cosine & \textbf{20} & \textbf{16.9} & \textbf{39.6} \\ \hline
\end{tabular}}
\caption{results with different beta schedulers}
\label{tab:beta}
\end{table}

\section{Conclusion}
\label{sec:conclusion}
In this paper we introduced a novel diffusion model for 
image captioning. We successfully applied the diffusion process to 
image captioning task with some modifications to fit into discrete
token generation while fusing image features. Results show that our model
can generate more diverse captions than previous autoregressive state-of-the-art
models and achieves comparable performance to the previous non-autoregressive
reseaches while our model with a simpler structure. We believe that
our model can be further improved with the community's exploration since
there are limited works on diffusion language generation, and we hope that 
our work can inspire more research in this field.

\section*{Limitations}
Our model is not applicable yet for its long sampling time, 
diffusion model requires a large number of time steps to keep its
markov property, fortunately, speeding up the sampling process is
a popular research topic in the field of diffusion model, accelerating
algorithms will benefit our model as well.

Another limitation is, though continuous diffusion can be applied,
there is still a gap between diffusion process and natural language meaning,
that is to say, token diffused with lower noise level does not mean it has
a closer meaning to the original token, while it is the case for images.
This limits the model to understand the diffusion process, and might be 
the reason why diffusion model for language generation training is not as stable
as training diffusion image generation models. We hope that future research can 
find a way to bridge this gap. 

\bibliography{acl2023.bib}
\bibliographystyle{acl_natbib}

\end{document}